%% file: main.tex
\documentclass[runningheads]{llncs}
\usepackage{graphicx}
\usepackage[utf8]{inputenc}
\usepackage{listings}
\usepackage{hyperref}
\usepackage{verbatim}
\usepackage{float}
\usepackage[ruled,vlined]{algorithm2e}
\usepackage{enumitem}
\usepackage{color}

\setlist[itemize]{noitemsep, topsep=0pt}
\setlist[enumerate]{noitemsep, topsep=0pt}

\definecolor{codegreen}{rgb}{0,0.6,0}
\definecolor{codegray}{rgb}{0.5,0.5,0.5}
\definecolor{codepurple}{rgb}{0.58,0,0.82}
\definecolor{backcolour}{rgb}{0.95,0.95,0.92}
 
\lstdefinestyle{mystyle}{
    backgroundcolor=\color{backcolour},   
    commentstyle=\color{codegreen},
    keywordstyle=\color{magenta},
    numberstyle=\tiny\color{codegray},
    stringstyle=\color{codepurple},
    basicstyle=\small,
    breakatwhitespace=false,         
    breaklines=true,                 
    captionpos=b,                    
    keepspaces=true,                 
    numbers=left,                    
    numbersep=5pt,                  
    showspaces=false,                
    showstringspaces=false,
    showtabs=false,                  
    tabsize=2
}

\begin{document}
\title{Testing Global Constraints}
%
%
\author{Aurélie Massart \and
Valentin Rombouts \and Pierre Schaus}
\authorrunning{Testing and debugging filtering of global constraints}
%
\institute{UCLouvain/ICTEAM Belgium}
\maketitle              
\begin{abstract}
Every Constraint Programming (CP) solver exposes a library of constraints for solving combinatorial problems. 
In order to be useful, CP solvers need to be bug-free. 
Therefore the testing of the solver is crucial to make developers and users confident.
We present a Java library allowing any JVM based solver
to test that the implementations of the individual constraints are correct.
The library can be used in a test suite executed in a continuous integration tool or it can also be used to discover minimalist instances violating some properties (arc-consistency, etc) in order to help the developer to identify the origin of the problem using standard debuggers.
\end{abstract}
\section{Introduction}

The filtering algorithms inside constraint programming solvers (\cite{choco,oscar,jacop,gecode} etc.)  are mainly tested using test suites implemented manually. 
Creating such unit tests is a significant workload for the developers and is also error prone.

The most elementary yet important test to achieve for a constraint is that no feasible solution is removed.
One can always implement a checker verifying the feasibility of the constraint when all the variables are bound. By comparing the number of solutions generated with both the checker and the tested filtering algorithm, one can be confident that no solution is removed.
This procedure can be repeated for many (small) instances (possibly randomly generated). 
Alternatively, one can compare with a decomposition of the constraint into (more) elementary ones. This latter approach can improve the coverage of the test suite.

Those unit tests verifying the non removal of feasible solutions do not verify other properties of constraints generally more difficult to test.
For instance, the domain-consistency property is rarely tested outside some hard-coded small test examples.

We introduce CPChecker as a tool to ease the solver developer’s life by automating the testing of properties of filtering algorithms. For instance, \textit{algorithm A should filter more than algorithm B} or \textit{Algorithm A should achieve arc or bound-consistency}, etc. 
The tool does not ensure that the tested filtering does not contain any bug - as it is impossible to test all possible input domains - but it can reveal the presence of one, if a test fails.
The large variety of input domains pseudo-randomly generated should make the user confident that the tool would allow to detect most of the bugs.

Many constraint implementations are stateful and maintain some reversible data structures. Indeed, global constraints' filtering algorithms often maintain an internal state in order to be more efficient than their decomposition.
This reversible state is also a frequent source of bugs. 
CPChecker includes the trail-based operations when testing constraints such that any bug due to the state management of the constraint also has a high chance to be detected.
CPChecker is generic and can be interfaced with any JVM trailed based solvers. 
CPChecker is able to generate detailed explanations by generating minimal domain examples on which the user's filtering has failed, if any.\\

\paragraph{Related work}
In \cite{meier1995debugging,coffrin17,lazaar2012cp}, the authors introduce tools to debug models. Some researches have also been done to help programmers while debugging codes for constraint programming \cite{StoreInspection}.
To the best of our knowledge, these tools, unlike CPChecker, do not focus on the filtering properties of individual constraints.

In the next sections, we first detail how to test static filtering algorithms before explaining the testing of stateful filtering algorithms for trailed based solvers. Finally we introduce how CPChecker can be integrated into a test suite.

\section{Testing Static Filtering Algorithms}

CPChecker is able to test any static filtering algorithm acting over integer domains. 
Therefore, the user needs to implement a function taking some domains (array of set of ints) as input and returning the filtered domains\footnote{Most of the code fragments presented are in Scala for the sake of conciseness but the library is compatible with any JVM-based language.}:

\begin{lstlisting}[language=scala,basicstyle=\small]
abstract class Filter {
 def filter(variables: Array[Set[Int]]): Array[Set[Int]]
}
\end{lstlisting}

CPChecker also needs a trusted filtering algorithm serving as reference with the same signature. 
The least effort for a user is to implement a checker for the constraint under the form of a predicate
that specifies the semantic of the constraint. 
For instance a checker for the constraint $\sum_i x_i=15$ can be defined as
\begin{lstlisting}[language=scala]
 def sumChecker(x: Array[Int]): Boolean = x.sum == 15
\end{lstlisting}

One can create with CPChecker an Arc/Bound-Z/Bound-D/Range Consistent filtering algorithm by providing in argument to the corresponding constructor the implementation of the checker. For instance

\begin{lstlisting}[language=scala,basicstyle=\small]
class ArcFiltering(checker: Array[Int] => Boolean) extends Filter
val trustedArcSumFiltering = new ArcFiltering(sumChecker)
\end{lstlisting}

This class implements the \texttt{filter} function as a trusted filtering algorithm reaching the arc consistency by 1) computing the Cartesian product of the domains, 2) filtering with the checker the non solutions and 3) creating the filtered domains as the the union of the values.
Similar filtering algorithms' (Bound-Z, Bound-D and Range) have been implemented from a checker.

Finally the \texttt{check} and \texttt{stronger} functions permit to respectively check that two compared filtering algorithms are the same or that the tested filtering is stronger than the trusted one.

\begin{lstlisting}[language=scala,basicstyle=\small,breaklines=true]
def check/stronger(trustedFiltering: Filter, testedFiltering: Filter) : Boolean
\end{lstlisting}

The testing involves the following steps: 
\begin{enumerate}
    \item Random Domains generation \footnote{A seed can be set to reproduce the same tests.}.
    \item Execution of the tested and trusted filtering algorithms (from CPChecker's filterings or another trusted one) to these random domains. 
    \item Comparison of the domains returned by the two filtering algorithms. 
\end{enumerate}
This process is repeated by default 100 times although all the parameters can be overridden for the creation of random domains, number of tests, etc.

\subsection{Generation of Random Test Instances}\label{section:gen}
In order to test a filtering implementation, CPChecker relies on a \textit{property based testing} library called \textit{ScalaCheck}\cite{scalaCheck}\footnote{Similar libraries exist for most programming languages, all inspired by QuickCheck for Haskell.}.
This library includes support for the creation of random generators and for launching multiple test cases given those. 
CPChecker also relies on the ability of \textit{ScalaCheck} of reducing the instance to discover a smaller test instance over which the error occurs.

\subsection{Example}
Here is an example for testing with CPChecker the arc-consistent \textit{AllDifferent} constraint's in \textit{\textbf{OscaR}} \cite{oscar} solver : 
\begin{lstlisting}[language=scala, basicstyle=\small]
object ACAllDiffTest extends App {
  def allDiffChecker(x: Array[Int]): Boolean = x.toSet.size == x.length
  val trustedACAllDiff: Filter = new ArcFiltering(allDiffChecker)
  val oscarACAllDiff: Filter = new Filter {
    override def filter(variables: Array[Set[Int]]): Array[Set[Int]] = {
      val cp: CPSolver = CPSolver()
      val vars = variables.map(x => CPIntVar(x)(cp))
      val constraint = new AllDiffAC(vars)
      try {
        cp.post(constraint)
      } catch {
        case _: Inconsistency => throw new NoSolutionException
      }
      vars.map(x => x.toArray.toSet)
    }
  }
  check(trustedACAllDiff, oscarACAllDiff)
}
\end{lstlisting}
The trusted filtering algorithm is created thanks to the \texttt{ArcFiltering} class at line 3. The checker for AllDifferent simply verifies that the union of the values in the array has a cardinality equal to the size of the array, as defined at line 2.
The tested filtering implements the \texttt{filter} function using \textit{\textbf{OscaR}}'s filtering. It first transforms the variables into \textit{\textbf{OscaR}}'s variables (line 7) then creates the constraint over them (line 8). It is then posted to the solver which filters the domains until fix-point before returning them.

\section{Testing stateful constraints}

Incremental Filtering Algorithms usually maintain some form of state in the constraints. 
It can for instance be reversible data-structures for trailed-based solvers.
CPChecker allows to test a stateful filtering algorithm by testing it during a search while checking the state restoration. 
In terms of implementation, the incremental \texttt{check} and \texttt{stronger} functions compare \texttt{FilterWithState} objects that must implement two functions. The \textit{setup} function reaches the fix-point while setting up the solver used for the search. The \textit{branchAndFilter} function applies a branching operation on the current state of the solver and reaches a new fix-point for the constraint. 
The branching operations represent standard branching constraints such as $=,\neq,<,>$ and the \texttt{push/pop} operations on the trail
allowing to implement the backtracking mechanism (see \cite{minicp} for further details on this mechanism).

\begin{lstlisting}[language=scala, basicstyle=\small]
abstract class FilterWithState {
  def setup(variables: Array[Set[Int]]): Array[Set[Int]]

  def branchAndFilter(branching: BranchOp): Array[Set[Int]]
}
\end{lstlisting}

The process of testing an incremental/stateful filtering algorithm is divided into four consecutive steps : 
\begin{enumerate}
    \item Domains generation
    \item Application of the \textit{setup} function of the tested and trusted filtering algorithms.
    \item Comparing the filtered domains returned at step 2.
    \item Execution of a number of fixed number dives as explained next based on the application of \textit{branchAndFilter} function.
\end{enumerate}
\subsection{Dives}
A dive is performed by successively interleaving a push of the state and a domain restriction operation. When a leaf is reached (no or one solution remaining) the dive is finished and a random number of states are popped to start a new dive as detailed in the algorithm \ref{algo:dives}.

\begin{algorithm}
\SetKwProg{Dives}{Dives}{}{}
\Dives{(root, trail, nbDives)}{
dives $\leftarrow$ 0\\
currentDomains $\leftarrow$ root \\
\While{dives $<$ nbDives}{
  \While{!currentDomains.isLeaf}{
    trail.push(currentDomains)\\
    restriction $\leftarrow$ new RandomRestrictDomain(currentDomains)\\
    currentDomains $\leftarrow$ branchAndFilter(currentDomains, restriction)\\
  }
  dives $\leftarrow$ dives + 1 \\
  \For{i $\leftarrow$ 1 to Random(1,trail.size-1)}{
  trail.pop()
  }
}
}
\caption{Algorithm performing dives}
\label{algo:dives}
\end{algorithm}
\newpage

\subsection{Illustration over an Example}
The next example illustrates CPChecker to test the \textit{\textbf{OscaR}}\cite{oscar}'s filtering for the constraint $\sum_i{x_i}=15$. It should reach Bound-Z consistency. 
\begin{lstlisting}[language=scala, basicstyle=\small]
object SumBCIncrTest extends App {

  def sumChecker(x: Array[Int]): Boolean = x.sum == 15
  val trusted = new IncrementalFiltering(new BoundZFiltering(sumChecker))
  val tested = new FilterWithState {
    val cp: CPSolver = CPSolver()
    var currentVars: Array[CPIntVar] = _
    
    override def branchAndFilter(branching: BranchOp): Array[Set[Int]] ={
      branching match {
        case _: Push => cp.pushState()
        case _: Pop => cp.pop()
        case r: RestrictDomain => try {
            r.op match {
              case "=" => cp.post(currentVars(r.index) === r.constant)
              ...}
          } catch { 
            case _: Exception => throw new NoSolutionException
          }
      }
      currentVars.map(x => x.toArray.toSet)
    }

    override def setup(variables: Array[Set[Int]]): Array[Set[Int]] = {
      currentVars = variables.map(x => CPIntVar(x))
      try {
        solver.post(sum(currentVars) === 15)
      } catch {
        case _: Exception => throw new NoSolutionException
      }
      currentVars.map(x => x.toArray.toSet)
    }
  }
  check(trusted, tested)
}
\end{lstlisting}
In this example, two \texttt{FilterWithState} are compared with the \texttt{check} function.

In CPChecker, the \texttt{IncrementalFiltering} class implements the  \\ \texttt{FilterWithState} abstract class for any \texttt{Filter} object. Therefore, the \\ \texttt{IncrementalFiltering} created with a \texttt{BoundZFiltering} object is used as the trusted filtering (line 4) which it-self relies on the very simple \texttt{sumChecker} function provided by the user and assumed to be bug-free.

\section{Custom Assertions}
To ease the integration into a JUnit like test suite, CPChecker has custom assertions extending the \textit{AssertJ}\cite{assertJ} library. The classes \texttt{FilterAssert} and \\ \texttt{FilterWithStateAssert} follow the conventions of the library with the \texttt{filterAs} and \texttt{weakerThan} functions to respectively test a filtering algorithm, as in the \texttt{check} and \texttt{stronger} functions. An example of assertion is:
\begin{lstlisting}[language=scala, basicstyle=\small]
assertThat(tested).filterAs(trusted1).weakerThan(trusted2)
\end{lstlisting}

\section{Code Source}
CPChecker's code source is publicly available in the \textit{Github} repository\footnotemark. This repository also contains several examples of usage of CPChecker with both \textit{Scala} solver and \textit{Java} solvers, namely  \textit{\textbf{OscaR}}\cite{oscar}, \textit{\textbf{Choco}}\cite{choco} and \textit{\textbf{Jacop}}\cite{jacop}. 
From those examples, \textit{CPChecker} detected that the arc consistent filtering of the \textit{Global Cardinality} constraint of \textit{\textbf{OscaR}} was not arc consistent for all the variables (the cardinality variables). 
This shows the genericity of \textit{CPChecker} and that it can be useful to test and debug filtering algorithms with only a small workload for the user.
Further details on the architecture and implementation of CPChecker can be found in the Master Thesis document available at the github repository\footnotemark[\value{footnote}].\footnotetext{https://github.com/vrombouts/Generic-checker-for-CP-Solver-s-constraints}

\section{Conclusion and Future Work}
This article presented CPChecker, a tool to test filtering algorithms implemented in any JVM-based programming language based on the JVM. Filtering algorithms are tested over domains randomly generated which is efficient to find unexpected bugs. Principally written in \textit{Scala}, CPChecker can be used to test simple and stateful filtering algorithms. 
It also contains its own assertions system to be directly integrated into test suites. 
As future work, we would like to integrate into CPChecker properties of scheduling filtering algorithms \cite{baptiste2001} 
such as edge-finder, not-first not-last, time-table consistency, energy filtering, etc. 
for testing the most recent implementation of scheduling algorithms \cite{gay2015time,dejemeppe2015unary,fahimi2014linear,vilim2007global,vilim2011timetable,tesch2016nearly}.

\input{bib.tex}

\end{document}

%% file: bib.tex



\pagebreak


\input{main.bbl}


%% file: main.bbl
\begin{thebibliography}{10}

\bibitem{choco}
Charles Prud'homme, Jean-Guillaume Fages, and Xavier Lorca.
\newblock {\em Choco Documentation}.
\newblock TASC - LS2N CNRS UMR 6241, COSLING S.A.S., 2017.

\bibitem{oscar}
{OscaR Team}.
\newblock {O}sca{R}: {S}cala in {O}{R}, 2012.
\newblock Available from \texttt{https://bitbucket.org/oscarlib/oscar}.

\bibitem{jacop}
{Krzysztof Kuchcinski, Radoslaw Szymanek and contributors}.
\newblock Jacop solver.
\newblock Available from
  \texttt{https://osolpro.atlassian.net/wiki/spaces/JACOP/}.

\bibitem{gecode}
{Gecode Team}.
\newblock Gecode: Generic constraint development environment, 2006.
\newblock Available from \texttt{http://www.gecode.org}.

\bibitem{meier1995debugging}
Micha Meier.
\newblock Debugging constraint programs.
\newblock In {\em International Conference on Principles and Practice of
  Constraint Programming}, pages 204--221. Springer, 1995.

\bibitem{coffrin17}
Peter J.~Stuckey Carleton~Coffrin, Siqi~Liu and Guido Tack.
\newblock Solution checking with minizinc.
\newblock In {\em ModeRef2017, The Sixteenth International Workshop on
  Constraint Modelling and Reformulation}, 2017.

\bibitem{lazaar2012cp}
Nadjib Lazaar, Arnaud Gotlieb, and Yahia Lebbah.
\newblock A cp framework for testing cp.
\newblock {\em Constraints}, 17(2):123--147, 2012.

\bibitem{StoreInspection}
Fr{\'e}d{\'e}ric Goualard and Fr{\'e}d{\'e}ric Benhamou.
\newblock {\em Debugging Constraint Programs by Store Inspection}, pages
  273--297.
\newblock Springer Berlin Heidelberg, Berlin, Heidelberg, 2000.

\bibitem{scalaCheck}
{ScalaCheck}.
\newblock \texttt{http://scalacheck.org}.

\bibitem{minicp}
Laurent Michel, Pierre Schaus, and Pascal~Van Hentenryck.
\newblock Minicp: A minimalist open-source solver to teach constraint
  programming.
\newblock {\em Technical Report}, 2018.

\bibitem{assertJ}
{AssertJ Library}.
\newblock \texttt{http://joel-costigliola.github.io/assertj/index.html}.

\bibitem{baptiste2001}
Philippe Baptiste, Claude Le~Pape, and Wim Nuijten.
\newblock {\em Constraint-based scheduling: applying constraint programming to
  scheduling problems}, volume~39.
\newblock Springer Science, 201.

\bibitem{gay2015time}
Steven Gay, Renaud Hartert, and Pierre Schaus.
\newblock Time-table disjunctive reasoning for the cumulative constraint.
\newblock In {\em International Conference on AI and OR Techniques in
  Constriant Programming for Combinatorial Optimization Problems}, pages
  157--172. Springer, 2015.

\bibitem{dejemeppe2015unary}
Cyrille Dejemeppe, Sascha Van~Cauwelaert, and Pierre Schaus.
\newblock The unary resource with transition times.
\newblock In {\em International conference on principles and practice of
  constraint programming}, pages 89--104. Springer, 2015.

\bibitem{fahimi2014linear}
Hamed Fahimi and Claude-Guy Quimper.
\newblock Linear-time filtering algorithms for the disjunctive constraint.
\newblock In {\em Proceedings of the Twenty-Eighth AAAI Conference on
  Artificial Intelligence}, pages 2637--2643. AAAI Press, 2014.

\bibitem{vilim2007global}
Petr Vil{\'\i}m.
\newblock Global constraints in scheduling.
\newblock 2007.

\bibitem{vilim2011timetable}
Petr Vil{\'\i}m.
\newblock Timetable edge finding filtering algorithm for discrete cumulative
  resources.
\newblock In {\em International Conference on AI and OR Techniques in
  Constriant Programming for Combinatorial Optimization Problems}, pages
  230--245. Springer, 2011.

\bibitem{tesch2016nearly}
Alexander Tesch.
\newblock A nearly exact propagation algorithm for energetic reasoning in o
  (n2logn).
\newblock In {\em International Conference on Principles and Practice of
  Constraint Programming}, pages 493--519. Springer, 2016.

\end{thebibliography}
